\pdfoutput=1
\documentclass[letterpaper]{article} % DO NOT CHANGE THIS
\usepackage[]{aaai23}  % DO NOT CHANGE THIS
\usepackage{times}  % DO NOT CHANGE THIS
\usepackage{helvet}  % DO NOT CHANGE THIS
\usepackage{courier}  % DO NOT CHANGE THIS
\usepackage[hyphens]{url}  % DO NOT CHANGE THIS
\usepackage{graphicx} % DO NOT CHANGE THIS
\usepackage{natbib}  % DO NOT CHANGE THIS AND DO NOT ADD ANY OPTIONS TO IT
\usepackage{caption} % DO NOT CHANGE THIS AND DO NOT ADD ANY OPTIONS TO IT
\usepackage{flushend}
\usepackage{algorithm}
\usepackage{algorithmic}

\usepackage{newfloat}
\usepackage{listings}
\DeclareCaptionStyle{ruled}{labelfont=normalfont,labelsep=colon,strut=off} % DO NOT CHANGE THIS
\floatstyle{ruled}
\newfloat{listing}{tb}{lst}{}
\floatname{listing}{Listing}
\title{Rehabilitating Homeless: Dataset and Key Insights\thanks{Supported by \textit ``Nochlezhka'' non-commercial organization for rehabilitation of the homeless people, see https://homeless.ru/en/}}
\author{
    Anna Bykova, \textsuperscript{\rm 1} 
    Nikolay Filippov, \textsuperscript{\rm 1} 
    Ivan P. Yamshchikov \textsuperscript{\rm 2, \rm3} 
}
\affiliations{
    \textsuperscript{\rm 1}LEYA Lab for Natural Language Processing, Higher School of Economics, Yandex, St. Petersburg, Russia\\
    \textsuperscript{\rm 2}Max Planck Institute for Mathematics in the Sciences, Leipzig, Germany\\
    \textsuperscript{\rm 3}CEMAPRE, University of Lisbon, Portugal\\
    ivan@yamshchikov.info
}

\begin{document}

\maketitle

\begin{abstract}
This paper presents a large anonymized dataset of homelessness alongside insights into the data-driven rehabilitation of homeless people. The dataset was gathered by a large non-profit organization working on rehabilitating the homeless for twenty years. This is the first dataset that we know of that contains rich information on thousands of homeless individuals seeking rehabilitation. We show how data analysis can help to make the rehabilitation of homeless people more effective and successful. Thus, we hope this paper alerts the data science community to the problem of homelessness. 
\end{abstract}

\section{Introduction}
\label{intro}
There is no universal definition of homelessness. In general, homelessness is the condition of lacking stable, safe, and adequate housing. In 2004, the United Nations sector of Economic and Social Affairs defined a homeless household as \textit{households without a shelter that would fall within the scope of living quarters. They carry their few possessions with them, sleeping in the streets, in doorways or on piers, or in any other space, on a more or less random basis.}\cite{UN2004}.

In 2009, at the United Nations Economic Commission for Europe Conference of European Statisticians (CES), held in Geneva, Switzerland, the Group of Experts on Population and Housing Censuses defined homelessness as follows:

In its Recommendations for the Censuses of Population and Housing, the CES identifies homeless people under two broad groups:
\begin{itemize}
\item[(a)] Primary homelessness (or rooflessness). This category includes persons living in the streets without a shelter that would fall within the scope of living quarters;
\item[(b)] Secondary homelessness. This category may include persons with no place of usual residence who frequently move between various types of accommodations (including dwellings, shelters, and institutions for the homeless or other living quarters). This category includes persons living in private dwellings but reporting ``no usual address'' on their census form.
\end{itemize}
The CES acknowledges that the above approach does not provide a full definition of the ``homeless'' \cite{UN2009}.

The problem of homelessness has attracted the attention of researchers since the 1980-s. Many researchers like \cite{Crane1998}, \cite{Avramov2001} or \cite{Collin1992} studied homelessness as primarily a housing problem. Others like \cite{Dear2014} or \cite{Wennig1991} studied homelessness among people with mental diseases attributing it to deinstitutionalizing state mental hospitals and reducing public expenditures on welfare. Many other aspects of homelessness were studied, providing various aggregate data on homelessness. There is much research on the key factors and areas that could be addressed to prevent homelessness; see, for example, \cite{Cunningham2006}, or \cite{Rukmana2010}. \cite{Oflaherty2004}, or \cite{Byrne2013} provide a meta-analysis of various previous research results and list variables negatively and positively affecting homelessness rates. For the recent meta-analysis, we address the reader to \cite{Rukmana2020}. This paper points out that many aspects of the homelessness problem and approaches to helping homeless people, as well as preventing homelessness, heavily depend on the country and local specifics.

In many developing countries, the homelessness problem is often practically addressed by charity and non-government organizations (NGOs) in parallel with the public sector. ``Nochlezhka'' charity organization is one of those helping homeless people in Moscow, Saint Petersburg, and other cities in Russia. Homeless people receive help in humanitarian and re-socialization projects. Humanitarian projects provide free overnight stay, hot food, access to a shower, hygiene products, and clothes. These services are provided to any homeless person, regardless of their intent to move into a permanent shelter, where they can get further assistance from social workers. Re-socialization projects are available for the homeless who decided to leave the street for the shelter and want to restore documents, find a job or find an independent residence. 

The employees of such organizations often face a necessity to make difficult and non-obvious decisions based on noisy, incomplete and conflicting data. Recently, data analytics and machine learning started to get wider adoption in assisting such decision-making. There are examples of machine learning applications to predict homelessness in Canada \cite{Arsenault2020}, \cite{VanBerlo2020}, to improve the housing system for homeless youth in the USA \cite{Chan2017} and others.

The history of ML applications in addressing homelessness is not long. However, a fairly large amount of data has already been gathered, see, for example, data.World portal \cite{DataWorld}. Although the data from the USA, Canada or UK is relatively big, we could not find many homelessness-related datasets from other countries. Also, the datasets available at the moment mostly contain aggregated data. The dataset presented in this paper has much more detailed (albeit anonymized) information.

We have been working together with ``Nochlezhka'' charity organization on studying the data collected so far in their databases. For the last ten years, ``Nochlezhka'' has stored its clients' data in its content management system (CMS). Every new homeless client, upon registry, is informed that their data are stored and could be used for research purposes. The original CMS data was stored in multiple tables. We cleaned, preprocessed, and anonymized the data creating a dataset available for further data analysis. In this paper, we also report results obtained with this dataset. These results provide valuable insights into the process of rehabilitation of the homeless. In particular, we demonstrate that certain factors predict the outcome of the person's rehabilitation activities with a reasonable level of certainty. We are sharing this dataset, together with the insights and conclusions made, with the data science community. We hope to attract more attention and data scientists to the problem of homelessness globally.

\section{Data}
\label{data}

The gathered dataset is one of the major contributions of this paper. In this section, we describe the preprocessing of original data and document the resulting dataset to facilitate its further use by other researchers interested in the problem of homelessness.

\subsection{Data Cleaning}
\label{data:preprpcessing:cleaning}
The data was downloaded from the CMS database directly. Some of the tables contained a lot of typos, mistakes, discrepancies, HTML tags, etc.; we cleaned this data. The biggest problem was typos and errors in names and dates of birth that resulted in misidentifying the same person (\textit{client} in database terminology) from different records as two or more people. We created person representation as a concatenation of full name and date of birth for each line in the clients' table. Then, we calculated the pairwise similarity using the Levenshtein distance between the obtained strings \cite{SeatGeekFuzzyWussy}. The pairs of strings with high similarity were further checked to identify if two table lines represent the same person. These checks were done manually due to some typos and errors in the original database. We also deleted all obviously fake birth dates, for example, 1001-01-07, thus obtaining a table of clients with a unique person at each line. There was still a small possibility that some person's information might be grouped with another under an erroneous name, but we estimate the probability of such error as extremely low. HTML tags and other service data were searched for and removed by regular expressions.

\subsection{Anonymizing}
\label{data:preprocessing:anonymizing}
There are two types of clients in the database: clients of humanitarian services and clients who participated in re-socialization projects. Once the client enters the re-socialization project, a numeric ID is assigned to the client in the database. All of their further activities are logged under this issued ID. The data on the clients using humanitarian services (shelter, free shower, etc.) was entered into the database the way it was received from the client. The fields included names, dates of birth, and other sensitive information. If a homeless person was enrolled in both humanitarian and re-socialization projects, we matched their ID. We assigned IDs from a separate pool for the names listed in the humanitarian database that had no matching ID. Thus we anonymized the data while leaving a possibility to distinguish between the clients who used humanitarian services only and those who entered re-socialization projects. There is no sensitive data in the dataset we are publishing -- all personal info is replaced with numeric IDs. We address ethical concerns of this decision in Section Ethics Concerns. Here we reiterate that the data is anonymized and published under explicit consent of the partner NGO and every client of the NGO projects.

\subsection{Obtaining Features from Database Tables}
\label{data:preprocessing:tables}

\subsubsection{Contract types}
\label{data:preprocessing:tables:types}

Upon requesting help from a social worker, a homeless person signs a contract between NGO's client and the NGO. In this contract, the client specifically lists the goal they want to achieve upon the contract completion. According to the previous experience of the social workers, such explicit commitment on the client's side improves the chances for successful rehabilitation. The dataset we publish with this paper is centered around predictions of \textit{contract} completion. Every contract is an entity within the CMS of ``Nochlezhka'' database representing some specific activity during re-socialization projects. We used the same term ``contract'' in the dataset. Depending on the client's goals, there are different types of a \textit{contract} with the organization like registration, document restoration, obtaining citizenship, etc. -- 43 types in total. A single person may have several contracts of different types, or several contracts of the same type in different time periods. We represent each occurrence of any contract type for each client as a separate line in the dataset.

\begin{figure*}
\includegraphics[width=0.86\textwidth]{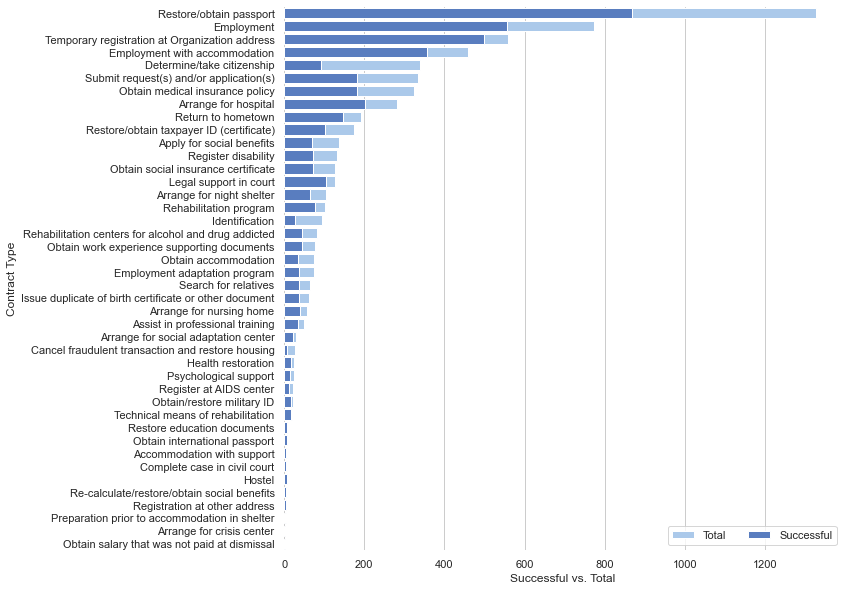}
\caption{Contract types distribution} \label{fig:types}
\end{figure*}

We have found that certain goals (such as restoring lost documents) are relatively easy to attain, while others (for example, rehabilitation for drug abuse) might have a lower success rate. To highlight this dependency on the type of a contract, we created a separate binary variable for each contract type -- that is, we one-hot encoded all 43 contract types. Figure~\ref{fig:types} illustrates the distribution of contract types in the dataset, with the share of successfully completed contracts of each type.

\subsubsection {Contract statuses}
\label{data:preprocessing:tables:statuses}

\begin{figure}
\includegraphics[width=0.5\textwidth]{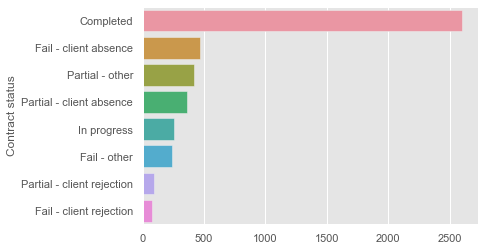}
\caption{Contract statuses distribution} \label{fig:statuses}
\end{figure}

There are eight \textit{contract statuses}: ``in progress'', ``fulfilled'', ``partially fulfilled due to client rejection'', ``partially fulfilled due to other reasons'', ``not fulfilled due to client rejection'', ``not fulfilled due to other reasons'', ``partially fulfilled due to client absence'', ``not fulfilled due to client absence''. We used the state of the database on the date 2021-07-31, containing records for over nine years, from 2012-07-11. Fig.~\ref{fig:statuses} shows the contract statuses distribution in the resulting dataset.

\subsubsection{Other features}
\label{data:preprocessing:tables:other}

We added data on the client's gender and age. Age in years was calculated separately based on the client's birth date, if available, and the slice date 2021-07-31. The database also contained additional information on the clients. Depending on the client, the features could include: marital status, dependants, profession, contacts with relatives, time in confinement, disability, alcohol abuse, drug abuse, and others -- the resulting table contained 28 columns of features derived from the database tables. %Table~\ref{tab:table_features} (see Appendices~\ref{appendix}) lists the full list of the resulting features in the published dataset.

\subsection{Obtaining features from comments}
\label{data:preprocessing:comments}
To obtain rich data on clients, we analyzed the fields with manual notes made by the social workers upon clients' registration. We found certain frequent remarks that could be relevant to the resulting dataset. These are remarks that social workers leave while working with a client. These comments are usually added upon the start of the \textit{contract} or during the interactions with the client along the rehabilitation process. Like other features, we publish the comment category tags for the comments added before the slice date 2021-07-31.

We removed punctuation, numbers, and other special symbols from the text of comments and extracted tokens using Natural Language Toolkit \cite{NLTK}. The resulting tokens were lemmatized using pymorphy2 \cite{pymorphy2}. The stopwords (prepositions, other auxiliary words) were removed from the resulting tokens list. The frequency dictionary was built from the received tokens list using the method proposed in \cite{rutermextract} to determine which keywords and topics appear in the manual comments more often. As a result of the analysis, 23 most frequent categories were selected. 
%(see Table~\ref{tab:table_comments}Appendices~\ref{appendix}).
Each of these categories was used as a tag. Some clients do not have any tags. A client may have more than one tag. The original texts of the comments made by the social workers are not included in the dataset. Thus, the dataset only includes the resulting frequent tags and does not include any client-specific personal or sensitive details. 

Finally, we obtained a unique anonymized \textit{client id} and corresponding \textit{tags} for every client. The most frequent tag is the tag \textit{comments\_relatives}. It occurs 2713 times in the dataset. The least frequent tag is  \textit{comments\_illiteracy} since only two clients were illiterate.

\subsection{Resulting dataset}
\label{data:dataset}
We merged the features obtained from database tables with the features obtained from comments by client ID to create the resulting dataset with a total of 6349 records and 51 features. Since contract type appeared to be an important feature worth further investigation, we also created an auxiliary dataset with all 43 contract type IDs one-hot encoded\footnote{Actually only 42 types were present, since one of the contract types was available in the database but did not occur on any record.}.

The dataset contains many NaN values -- not every record contains every feature. There are cases where some feature values for specific records are unknown. Therefore, additional preprocessing will be necessary for the machine learning models that can not work with NaN values.

We publish raw anonymized data on clients, contracts, and humanitarian projects that may help the data science community try different hypotheses, approaches, and models in studying the dataset. The data is cleaned from errors, duplicates, and discrepancies and translated into English.

%Table~\ref{tab:table_dataset} (see Appendices~\ref{appendix}) lists the columns of the resulting dataset.
Table \ref{tab:desc_dataset} shows some statistics describing the resulting dataset.

\begin{table}
\begin{center}
\begin{tabular}{|p{0.3\textwidth}|p{0.1\textwidth}|}
\hline
\bf{Description}              & \bf{Quantity}\\
\hline
Total records (different contract items) & 6349\\
\hline
Different contracts & 3589\\
\hline
Different clients & 2754\\
\hline
Men & 2239\\
\hline
Women & 515\\
\hline
Age 0-20 & 7\\
\hline
Age 21-40 & 789\\
\hline
Age 41-60 & 1467\\
\hline
Age 61-80 & 476\\
\hline
Age 80+ & 14\\
\hline
Contract type 8 Temp. registration & 558\\
\hline
Contract type 17 Citizenship & 339\\
\hline
Contract type 20 Cancel fraudulent transaction, return housing & 26\\
\hline
Comments: pregnancy & 11 clients\\
\hline
Contract type 25 Military Service Card & 21\\
\hline
\end{tabular}
\caption{The description of the dataset. Contracts period of time from 2012 to 2021 (9 years).}\label{tab:desc_dataset}
\end{center}
\end{table}

\section{Models}
\label{models}
To see if the obtained dataset could provide meaningful insights into the rehabilitation of the homeless, we have trained several models to predict the probability of rehabilitation success using available data about the client.

\subsection{Target variable}
\label{models:target}

Since in the experiment described here, we make a prediction based on client data, let us consider ``\textit{fulfilled}'' status (see ``Contract statuses'' subsection above) as a positive result, while statuses other than ``\textit{fulfilled}'' and directly related to the client (``partially fulfilled due to client rejection'', ``not fulfilled due to client rejection'', ``partially fulfilled due to client absence'', ``not fulfilled due to client absence'') as negative results. We considered a total of 6349 contracts, while 923 contracts with statuses: ``in progress'', ``partially fulfilled due to other reasons'', and ``not fulfilled due to other reasons'' were not included in our consideration in the subset for the particular experiment (although this data is still available in the dataset we publish for further experiments and research). Thus, we simplified the task down to binary classification with two values of the target variable: 1 - ``contract completed successfully'' and 0 - ``contract not completed due to the client's behavior''.
%Fig.~\ref{fig:target} shows the final target variable distribution in the dataset.

%\begin{figure}
%\includegraphics[width=0.5\textwidth]{target.jpg}
%\caption{Target variable distribution} \label{fig:target}
%\end{figure}

\subsection{Models and Experiments}
\label{models:results}

The logistic regression model was used as a baseline. Along with logistic regression, we experimented with support vector machine (SVM), decision tree classifier, random forest classifier, k nearest neighbors algorithm, fully connected neural network (FCNN), CatBoost classifier, and XGBoost classifier. We used the train-test-split module from Python sklearn library to split the dataset into train and validation subsets. 80~\% of the records were used in the training set, 20~\% in the validation set. We kept the positive-negative ratio in both training and validation sets as in the original set (``stratify'' property of train-test-split). Then we balanced the data in the training set, sampling an equal share of successful and unsuccessful contracts to train the model. Two methods (random undersampling and random oversampling) were used to balance the training subset. The validation subset was not modified and kept the original positive-negative ratio. We have then trained several models on the train set. Table \ref{tab:model_scores} summarizes the results of the trained models on the validation subset for two different balancing techniques. Since the validation set is imbalanced, we use the weighted average of F1-score as the model quality measure.

\begin{table}
\begin{center}
\begin{tabular}{|l|l|l|}
\hline
\bf{Model}              & \vtop{\hbox{\strut \bf{F1 with}}\hbox{\strut \bf{Undersampling}}}  & \vtop{\hbox{\strut \bf{F1 with}}\hbox{\strut \bf{Oversampling}}}\\
\hline
Logistic Regression     & 0.64 & 0.65\\
\hline
SVM                     & 0.64 & 0.64 \\
\hline
Decision Tree           & 0.73 & 0.76 \\
\hline
k Nearest Neighbours    & 0.65 & 0.65 \\
\hline
FCNN                    & 0.69 & 0.74 \\
\hline
CatBoost                & 0.76 & 0.76 \\
\hline
Random Forest           & 0.78 & 0.81 \\
\hline
XGBoost                 & \bf{0.80} & \bf{0.85} \\
\hline
\end{tabular}
\caption{Resulting performance of the tested models (weighted average of F1-score on the balanced training set with different balancing techniques).}\label{tab:model_scores}
\end{center}
\end{table}

Since the majority of the variables in the dataset are categoric, it is not surprising that CatBoost \cite{dorogush2018catboost} and XGBoost \cite{chen2016xgboost} perform well on the provided data. These algorithms were explicitly designed to work with categoric variables. However, what is surprising is the relatively high F-1 score (as well as a comparable accuracy) that the models demonstrate on the test. Such information could be helpful for social workers since ``Nochlezhka'' is a non-profit organization that has to function under considerable financial constraints. Using non-balanced data and adjusting for the discrepancies in the size of different data classes, one can obtain higher scores\footnote{The code of experiments is available at https://github.com/LEYADEV/homeless}. 

It is also important to note that a high F-1 score also validates the obtained dataset. It shows that the provided data contains meaningful information on the nature of homelessness available on an individual's level, which makes the provided dataset unique for the field. In the next section, we discuss further insights that could be obtained with the provided dataset.

\section{Discussion}
\label{discussion}

\begin{figure*}
\includegraphics[width=0.9\textwidth]{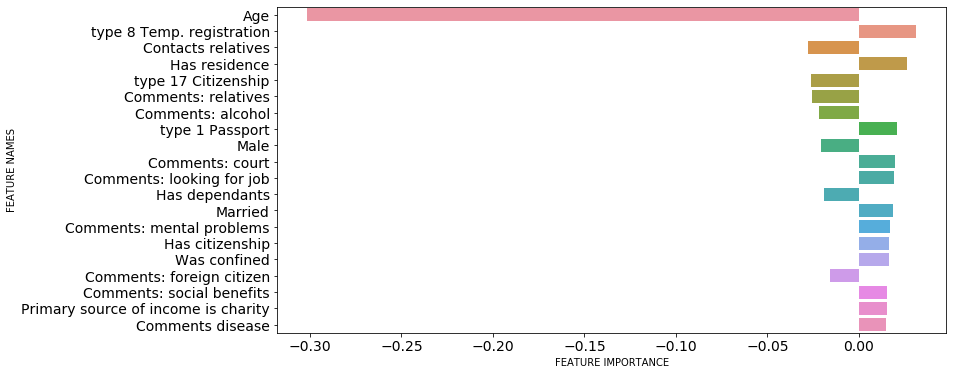}
\caption{Top 20 most important features: Decision Tree Classifier with Random Oversampling} \label{fig:dt_top20}
\end{figure*}

\begin{figure*}
\includegraphics[width=0.9\textwidth]{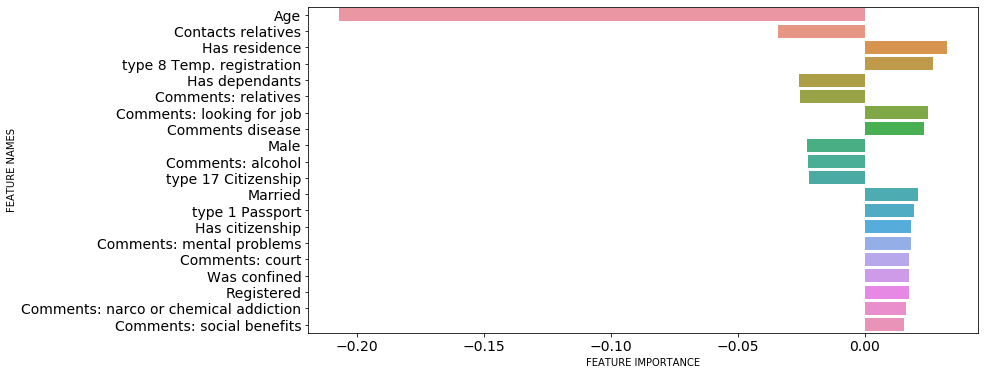}
\caption{Top 20 most important features: Random Forest Classifier with Random Oversampling} \label{fig:rf_top20}
\end{figure*}

\begin{figure*}
\includegraphics[width=0.9\textwidth]{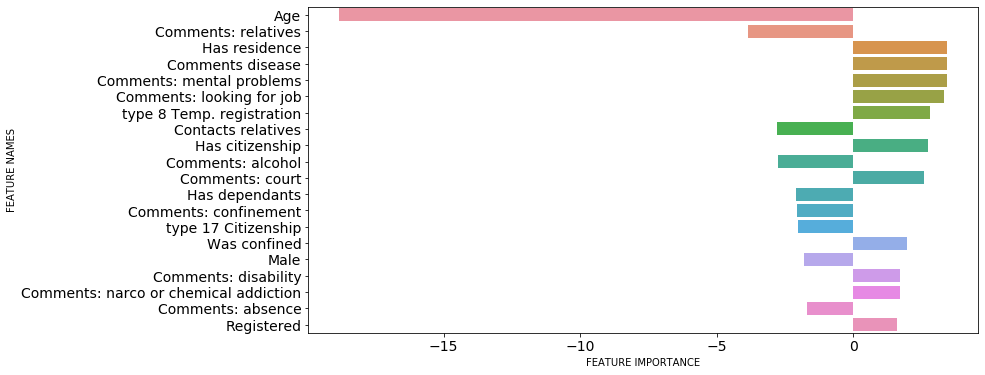}
\caption{Top 20 most important features: CatBoost Classifier with Random Oversampling} \label{fig:cb_top20}
\end{figure*}

\begin{figure*}
\includegraphics[width=0.9\textwidth]{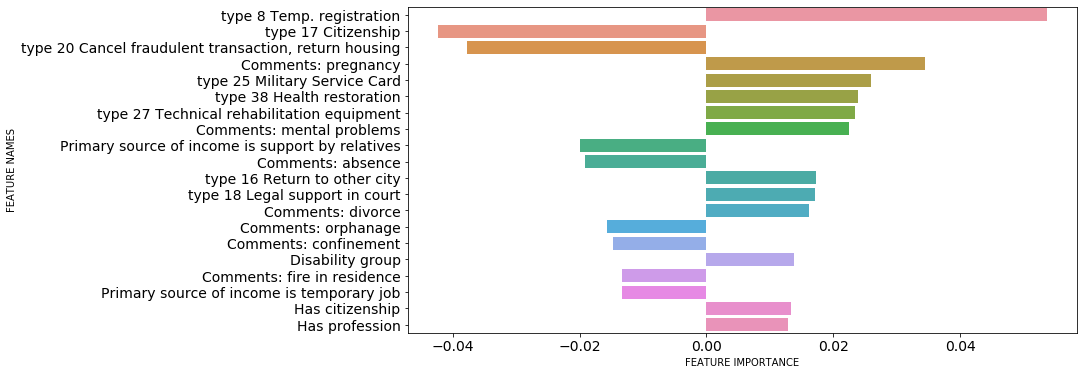}
\caption{Top 20 most important features: XGBoost Classifier with Random Oversampling} \label{fig:xgb_top20}
\end{figure*}

%\begin{figure}%[ht!]
%\includegraphics[width=0.5\textwidth]{importance_.jpg}
%\caption{Feature importance for XGBoost} %\label{fig:importance}
%\end{figure}

Figures~\ref{fig:dt_top20}--\ref{fig:xgb_top20} summarize the importance of various features that, according to different models, influence the contract success the most, either in ``successful'' or ``unsuccessful'' direction. All models for which the feature importance is illustrated, are tree-based. Therefore, they have similar criteria for feature importance measure calculation. In tree-based models, the importance is generally calculated for a single decision tree by the amount each feature split point improves the performance measure, weighted by the number of observations the node is responsible for. All four models above have ``feature\_importance\_'' method that was used to calculate the feature importance absolute value. The sign was taken from the correlation table and represented the sign of the correlation between a specific feature and the target variable. One sees that, although not all the models have the same features on their top-20 importance list, some features appear in most, if not all, top lists. For example, one sees that the essential feature is a registered absence of the client. This typically means the client missed one or several appointments with the social worker. Other high-impact factors include age, contact with relatives, diseases, and disability. We have discussed the results with social workers. While doing so, we made it clear to the social workers that feature importance does not, in any case, mean any cause-and-effect relationship between a specific feature and the result; just the extent to which the model regards the feature as essential for the result prediction. The feedback was that along with some expected results, there are certain surprising aspects.

The following features have the highest positive impact on XGBoost model that has shown the best results:

\begin{itemize}
\item the person seeks registration for the shelter provided by ``Nochlezhka'';
\item the person or the person's spouse is pregnant;
\item the person seeks to restore documents -- namely, the military service card or a passport;
\item the person seeks medical attention;
\item the person needs technical rehabilitation equipment;
\item the person or someone close to the person has mental problems;
\item the person wants to return to hometown, where they might have a place to live;
\item the person seeks legal support in court;
\item the person is divorced;
\item the person has disabilities;
\item the person is the citizen of Russia;
\item the person has a profession.
\end{itemize}

The following features have the highest negative impact, in accordance with the same XGBoost model:

\begin{itemize}
\item the person seeks to obtain citizenship of Russia;
\item the person seeks to return their home in court since the home was lost due to a fraudulent behavior of the third party;
\item primary source of the person's income is support from relatives;
\item the person was absent on one or more of the scheduled meetings;
\item the person is orphanage graduate;
\item the person mentions doing time in confinement;
\item the person's residence suffered from fire;
\item primary source of the person's main income was a temporary job.
\end{itemize}

While some of the factors on the list above are not surprising, others drew the attention of the social workers. First, success heavily depends on the person's goals: some goals are easier to achieve. Some of the differences are not obvious yet matter. For example, seeking a job that provides accommodation (e.g., a nightguard or a municipal caretaker)  contributes to success, while pursuing an adaptation program that claims to help with future employment (although not on the list above but still relatively significant feature) contributes negatively. This insight gives a clear and helpful outcome: the currently implemented employment program has room for further improvement. For example, ``Nochlezhka'' might partner with employers who could provide accommodation for the person they hire and develop programs that could further simplify the search for employment that also provides lodging. It is well known that drug abuse and history of confinement negatively affect rehabilitation outcomes. It is also relatively well known that once gathering becomes a person's main activity, this usually signifies that the person has been ``chronically'' homeless. The social workers know from experience that the longer someone is on the street, the harder it is to achieve rehabilitation.

The real-life use cases for predictions and feature importance evaluations made by machine learning models do not imply ``only concentrate on those contracts that are likely to succeed, ignore the rest'' strategy. The ML-powered methods mostly highlight the aspects social workers need to pay more attention to if they want to achieve a successful result with a specific client. Another useful application is backing up the organization's position in discussions with state officials: the arguments based on knowledge from social workers' experiences become much more convincing when confirmed by statistics and machine learning.

We believe that further analysis of the dataset could provide deeper insights into the challenges of rehabilitation of the homeless.

\section{Ethics Concerns}
\label{sec:ethics}

Homelessness is an extremely sensitive issue. Unfortunately, to this day, social stigma is attached not only to homelessness per se but also to other issues that tend to align with the problem of losing a home: substance abuse, family violence, and mental illness are to name a few. This stigma makes quantitative research on homelessness very challenging. Most of the available datasets only include macro-level statistics on homeless people and rarely allow precise analysis of individuals. Even data on various social and demographic cohorts are rare. However, homelessness is a multifaceted issue that, in our opinion, demands rich, multidimensional datasets describing the problem on the level of individuals since every person has a unique story and a unique set of factors that affect the rehabilitation process. Working on this contribution, we tried to balance these two conflicting factors to the best of our capabilities.

Several years ago, our partnering non-government charity organization entertained the thought that detailed research of the collected data might benefit other organizations that work on rehabilitating the homeless. The social worker informed the clients that their data might help other homeless people. Upon registering a contract with the NGO, every client was asked to sign an informed consent that their data could be processed, depersonalized, and published under several constraints that we followed. We have double-checked with the NGO's social workers and volunteers the scale of potential risks for their clients upon deanonymization. They all agree that such risks are minimal since the NGO encourages their clients to share their stories and thus combat social stigmas that surround homelessness. For example, most of the clients, even in the middle of the rehabilitation process, agree to talk about their past to the journalists. The clients understand the importance of higher visibility for homeless people and try to do their share in educating society about the issue. At the same time, we did our best to preprocess the data so that we estimate the risk of deanonymization as low. Those two factors combined give us ground to conclude that the benefits of a detailed anonymized public dataset of homelessness spanning several years and thousands of clients outweigh the potential risks such publication might bring for specific individuals. 

\section{Conclusion}
\label{conclusion}

This paper presents a novel large dataset of homelessness. The dataset consists of more than six thousand records. The data is anonymized yet provides many details on each client's level that used services of the rehabilitation organization. These details shed light on various factors that affect the success of rehabilitation. We publish the dataset for further research. We demonstrate that one could predict the probability of rehabilitation with an accuracy of up to 80~\% and more. The results obtained within the project are used by the non-profit organization that rehabilitates homeless people. We hope that the publication of these rich and socially vital data could attract the attention of the data science community to the problem of homelessness and provide new effective solutions that could help the homeless worldwide.

\appendix
\section{Appendices}
\label{appendix}

Appendix to this paper is available at https://github.com/LEYADEV/homeless. It contains tables with detailed description of category tags obtained from dataset tables, category tags obtained from comments, and all dataset columns, along with NaN statistics.

\section*{Acknowledgements}
\label{acknowledgements}

This work was supported by the grant for research centers in the field of AI provided by the Analytical Center for the Government of the Russian Federation (ACRF) in accordance with the agreement on the provision of subsidies (identifier of the agreement 000000D730321P5Q0002) and the agreement with HSE University No. 70-2021-00139.

We thank Daniil Alexandrov, Danil Kramorov and the whole team of \textit ``Nochlezhka'' for their help, good will and readiness to support us in this project.

\bibliography{bib.bib}

\begin{thebibliography}{22}
\providecommand{\natexlab}[1]{#1}

\bibitem[{Arsenault(2020)}]{Arsenault2020}
Arsenault, C. 2020.
\newblock Using AI, Canadian city predicts who might become homeless.
\newblock \emph{Thomson Reuters Foundation News}, October.

\bibitem[{Avramov(2001)}]{Avramov2001}
Avramov, D. 2001.
\newblock The changing face of homelessness in Europe.
\newblock In \emph{International perspective on homelessness}, 3--38. Greenwood
  Press.

\bibitem[{Byrne et~al.(2013)Byrne, Munley, Fargo, Montgomery, and
  Culhane}]{Byrne2013}
Byrne, T.; Munley, E.; Fargo, J.~D.; Montgomery, A.~E.; and Culhane, D.~P.
  2013.
\newblock New perspectives on community-level determinants of homelessness.
\newblock \emph{Journal of Urban Affairs}, 35(5): 607--625.

\bibitem[{Chan et~al.(2017)Chan, Rice, Vayanos, Tambe, and Morton}]{Chan2017}
Chan, H.; Rice, E.; Vayanos, P.; Tambe, M.; and Morton, M. 2017.
\newblock Evidence From the Past: AI Decision Aids to Improve Housing Systems
  for Homeless Youth.
\newblock In \emph{Proc. of AAAI Fall Symposium Series on Cognitive Assistance
  in Government and Public Sector Applications}.

\bibitem[{Chen and Guestrin(2016)}]{chen2016xgboost}
Chen, T.; and Guestrin, C. 2016.
\newblock Xgboost: A scalable tree boosting system.
\newblock In \emph{Proceedings of the 22nd acm sigkdd international conference
  on knowledge discovery and data mining}, 785--794.

\bibitem[{Collin(1992)}]{Collin1992}
Collin, R.~W. 1992.
\newblock Homelessness in the United States: 1980–-1990.
\newblock \emph{Journal of Planning Literature}, 7(1): 22--37.

\bibitem[{Crane and Takashi(1998)}]{Crane1998}
Crane, R.; and Takashi, L.~M. 1998.
\newblock Who are the suburban homeless and what do they want? An empirical
  study of the demand for public services.
\newblock \emph{Journal of Planning Education and Research}, 18(1): 35--48.

\bibitem[{Cunningham, Schmitt, and Henry(2006)}]{Cunningham2006}
Cunningham, M.; Schmitt, E.; and Henry, M., eds. 2006.
\newblock \emph{A new vision: What is in community plans to end homelessness}.
\newblock National Alliance to End Homelessness.

\bibitem[{Data.World(2022)}]{DataWorld}
Data.World. 2022.
\newblock Homelessness datasets.
\newblock https://data.world/datasets/homelessness.
\newblock Accessed: 2022-04-01.

\bibitem[{Dear and Wolch(2014)}]{Dear2014}
Dear, M.~J.; and Wolch, J.~R., eds. 2014.
\newblock \emph{Landscapes of despair: From deinstitutionalization to
  homelessness}.
\newblock Princeton University Press.

\bibitem[{Dorogush, Ershov, and Gulin(2018)}]{dorogush2018catboost}
Dorogush, A.~V.; Ershov, V.; and Gulin, A. 2018.
\newblock CatBoost: gradient boosting with categorical features support.
\newblock arXiv:1810.11363.

\bibitem[{https://www.nltk.org/team.html(2021)}]{NLTK}
https://www.nltk.org/team.html. 2021.
\newblock Natural Language Toolkit.
\newblock https://www.nltk.org/.
\newblock Accessed: 2021-12-01.

\bibitem[{Korobov(2020)}]{pymorphy2}
Korobov. 2020.
\newblock Morphological analyzer.
\newblock https://github.com/kmike/pymorphy2.
\newblock Accessed: 2021-12-01.

\bibitem[{O'Flaherty(2004)}]{Oflaherty2004}
O'Flaherty, B. 2004.
\newblock Wrong person and wrong place: For homelessness, the conjunction is
  what matters.
\newblock \emph{Journal of Housing Economics}, 13(1): 1--15.

\bibitem[{Rukmana(2010)}]{Rukmana2010}
Rukmana, D. 2010.
\newblock Gender differences in the residential origins of the homeless:
  Identification of areas with high risk of homelessness.
\newblock \emph{Planning, Practice and Research}, 25(1): 95--116.

\bibitem[{Rukmana(2020)}]{Rukmana2020}
Rukmana, D. 2020.
\newblock The Causes of Homelessness and the Characteristics Associated With
  High Risk of Homelessness: A Review of Intercity and Intracity Homelessness
  Data.
\newblock \emph{Housing Policy Debate}.

\bibitem[{SeatGeek(2022)}]{SeatGeekFuzzyWussy}
SeatGeek. 2022.
\newblock FuzzyWuzzy Python package.
\newblock https://github.com/seatgeek/fuzzywuzzy.
\newblock Accessed: 2021-12-01.

\bibitem[{Shevchenko(2018)}]{rutermextract}
Shevchenko, I. 2018.
\newblock rutermextract.
\newblock https://github.com/igor-shevchenko/rutermextract.
\newblock Accessed: 2021-12-01.

\bibitem[{UN(2004)}]{UN2004}
UN. 2004.
\newblock United Nations Demographic Yearbook review: National reporting of
  household characteristics, living arrangements and homeless households :
  Implications for international recommendations.
\newblock Department of Economic and Social Affairs, Statistics Division,
  Demographic and Social Statistics Branch.

\bibitem[{UN(2009)}]{UN2009}
UN. 2009.
\newblock Enumeration of Homeless People.
\newblock Economic and Social Council, Economic Commission for Europe
  Conference of European Statisticians, Group of Experts on Population and
  Housing Censuses, Twelfth Meeting.

\bibitem[{VanBerlo et~al.(2020)VanBerlo, Ross, Rivard, and
  Booker}]{VanBerlo2020}
VanBerlo, B.; Ross, M. A.~S.; Rivard, J.; and Booker, R. 2020.
\newblock Interpretable Machine Learning Approaches to Prediction of Chronic
  Homelessness.
\newblock arXiv:2009.09072.

\bibitem[{Wennig(1991)}]{Wennig1991}
Wennig, M.~V. 1991.
\newblock The homeless mentally III.
\newblock \emph{Journal of Planning Literature}, 5(3): 307--314.

\end{thebibliography}

\end{document}